\documentclass[letterpaper]{article} 
\usepackage[]{aaai23}  
\usepackage{times}  
\usepackage{helvet}  
\usepackage{courier}  
\usepackage[hyphens]{url}  
\usepackage{graphicx} 
\usepackage{natbib}  
\usepackage{caption} 
\usepackage{algorithm}
\usepackage{algorithmic}

\usepackage{amsmath,amsfonts,bm}

\def\eqref#1{equation~\ref{#1}}

\def\1{\bm{1}}

\def\vx{{\bm{x}}}
\def\vy{{\bm{y}}}

\DeclareMathAlphabet{\mathsfit}{\encodingdefault}{\sfdefault}{m}{sl}
\SetMathAlphabet{\mathsfit}{bold}{\encodingdefault}{\sfdefault}{bx}{n}

\usepackage{tikz}
\usepackage{comment}
\usepackage{amsmath,amssymb} 
\usepackage{color}
\usepackage{multirow}
\usepackage{tabularx, booktabs} 

\newcommand{\ie}{\emph{i.e.}}
\newcommand{\eg}{\emph{e.g.}}

\usepackage{newfloat}
\usepackage{listings}
\DeclareCaptionStyle{ruled}{labelfont=normalfont,labelsep=colon,strut=off} 
\floatstyle{ruled}
\newfloat{listing}{tb}{lst}{}
\floatname{listing}{Listing}
\title{Adversarial Style Augmentation for Domain Generalization}
\author{
    Yabin Zhang$^1$,
    Bin Deng$^2$,
    Ruihuang Li$^1$,
    Kui Jia$^2$,
    Lei Zhang$^1$
}
\affiliations{
    \textsuperscript{\rm 1} The Hong Kong Polytechnic University \textsuperscript{\rm 2} South China University of Technology \\

    csybzhang@comp.polyu.edu.hk
}

\usepackage{bibentry}

\begin{document}
	\newcolumntype{L}[1]{>{\raggedright\arraybackslash}p{#1}}
	\newcolumntype{C}[1]{>{\centering\arraybackslash}p{#1}}
	\newcolumntype{R}[1]{>{\raggedleft\arraybackslash}p{#1}}
	\maketitle

	\begin{abstract}
		It is well-known that the performance of well-trained deep neural networks may degrade significantly when they are applied to data with even slightly shifted distributions. 
		Recent studies have shown that introducing certain perturbation on feature statistics (\eg, mean and standard deviation) during training can enhance the cross-domain generalization ability. Existing methods typically conduct such perturbation by utilizing the feature statistics within a mini-batch, limiting their representation capability. Inspired by the domain generalization objective, we introduce a novel Adversarial Style Augmentation (ASA) method, which explores broader style spaces by generating more effective statistics perturbation via adversarial training. 
		Specifically, we first search for the most sensitive direction and intensity for statistics perturbation by maximizing the task loss. By updating the model against the adversarial statistics perturbation during training, we allow the model to explore the worst-case domain and hence improve its generalization performance. To facilitate the application of ASA, we design a simple yet effective module, namely AdvStyle, which instantiates the ASA method in a plug-and-play manner. We justify the efficacy of AdvStyle on tasks of cross-domain classification and instance retrieval. It achieves higher mean accuracy and lower performance fluctuation.
		Especially, our method significantly outperforms its competitors on the PACS dataset under the single source generalization setting, \eg, boosting the classification accuracy from 61.2\% to 67.1\% with a ResNet50 backbone. Our code will be available at \url{https://github.com/YBZh/AdvStyle}.
		
		
	\end{abstract}

	\section{Introduction} \label{Sec:introduction}
	
	Deep neural networks (DNN) have exhibited impressive performance on many vision tasks, especially when the training and test data follow the same distribution, \ie, the so-called assumption of independent and identical distribution (IID). Unfortunately, the IID assumption may not hold in practical applications. For instance, detection models trained on samples collected in sunny days may be applied to data collected in bad weather (\eg, rainy days). In such cross-domain application scenarios, the DNN models may suffer from significant performance degradation. 
	To solve this issue, domain generalization (DG) has been introduced to improve the generalization performance of DNN models to unseen test domains, and many DG methods have been developed recently \cite{motiian2017unified,li2018domain,gong2019dlow,zhao2020domain,li2018learning,du2020learning,volpi2018generalizing,yue2019domain,zhou2020learning,geirhos2018imagenet,nam2021reducing,zhou2021domain}.

	Generally speaking, DG aims to learn a model from the source domain(s) while ensuring that the learned model could perform well to any unseen test domains. 	
	The DG problem can be conventionally solved by learning domain invariant features \cite{motiian2017unified,li2018domain,gong2019dlow,zhao2020domain}, employing meta-learning strategies \cite{li2018learning,du2020learning}, performing data augmentations \cite{volpi2018generalizing,yue2019domain,zhou2020learning,geirhos2018imagenet,nam2021reducing,zhou2021domain}, and so on \cite{wang2021generalizing,zhou2021survey}. 
	Recently, performing style augmentation in feature space by conducting feature statistics perturbation has attracted increasing attention due to its simplicity and efficacy \cite{nuriel2021permuted,zhou2021domain,zhang2022exact,li2022uncertainty}. Specifically, it has been observed in the application of style transfer \cite{huang2017arbitrary,dumoulin2016learned} that feature statistics (\eg, mean and standard deviation) characterize the style information, and changing such statistics of image features will result in style-changed but content-preserved output images. Inspired by this observation, researchers have proposed to perform feature statistics perturbation to introduce style-augmented samples \cite{nuriel2021permuted,zhou2021domain,zhang2022exact,li2022uncertainty,chen2022maxstyle,zhong2022adversarial}. By expanding the training data with these style-augmented samples, improved generalization performance of DNN models has been observed \cite{nuriel2021permuted,zhou2021domain,zhang2022exact,li2022uncertainty}.
	
	In such a statistics perturbation-based DG framework, the perturbation strategy is of vital importance.  Nuriel \emph{et al.} \cite{nuriel2021permuted} and Zhou \emph{et al.} \cite{zhou2021domain} respectively introduced the statistics perturbation by randomly swapping or linearly interpolating statistics of two instances within a mini-batch. Li \emph{et al.} \cite{li2022uncertainty} sampled the feature statistics perturbation from Gaussian distributions, the means and standard deviations of which are estimated from the current mini-batch.  Although improved performance has been achieved, existing methods, unfortunately, constrain the representation space of feature statistics perturbation within that of the current mini-batch, limiting the diversity of style augmentations. 
	
	To explore a broader style space beyond that spanned by batch statistics, we propose to generate more diverse statistics perturbation.
	Similar to \cite{li2022uncertainty}, we model the feature statistics (\ie, mean and standard deviation) as Gaussian distributions and utilize the vanilla feature statistics as the means of these Gaussians. In \cite{li2022uncertainty}, the standard deviations of these Gaussians are estimated from the current mini-batch. In contrast, inspired by the DG objective (see Equ. (\ref{Equ:dg_objective})), we model these standard deviations as learnable parameters and optimize them via adversarial training, resulting in the Adversarial Style Augmentation (ASA) method. Specifically, by maximizing the task loss w.r.t. the learnable standard deviations, we approach the most sensitive perturbation direction and intensity (\ie, the worst-case domain). Meanwhile, by minimizing the task loss w.r.t. the vanilla task model, we update the model against the worst-case domain perturbation. The model is expected to generalize well to tough unseen test domains if it could generalize to the worst-case domain.

\begin{figure}[tb]
		\begin{center}
			\includegraphics[width=0.99\linewidth]{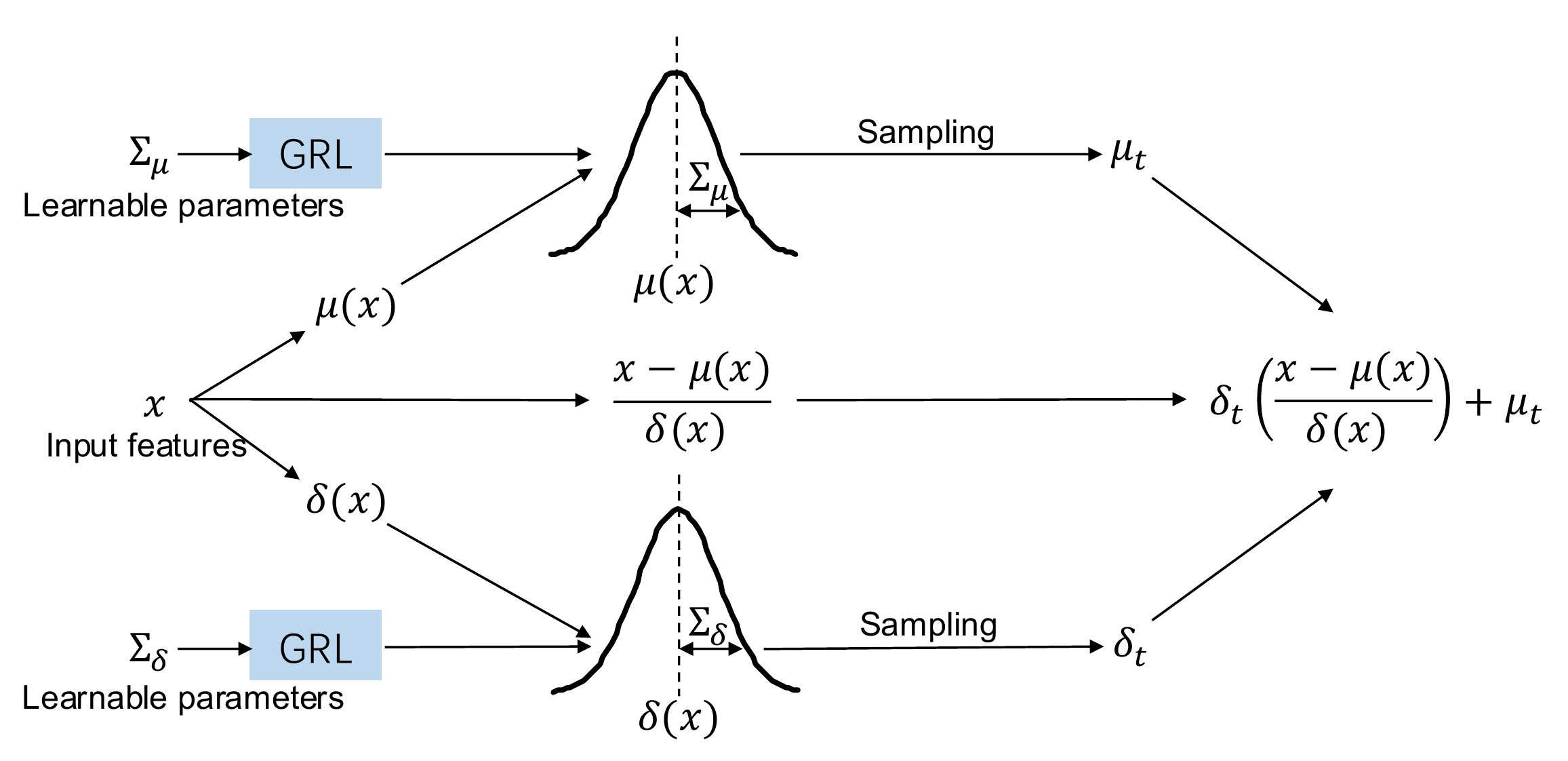}
		\end{center}
		\vspace{-0.3cm}
		\caption{An illustration of the AdvStyle module, where 'GRL' is the gradient reverse layer \cite{ganin2015unsupervised,long2018conditional}.
		} \label{Fig:advstyle}
	\end{figure}

	The proposed ASA method could be directly implemented by conducting the maximization and minimization steps iteratively (please refer to Fig. \ref{Fig:two_step} for more details). As illustrated in Sec. \ref{Subsec:ablation}, such an iterative optimization strategy could lead to outstanding performance. To facilitate the application of ASA in practice, we take a step further and propose a simple yet effective module, namely AdvStyle, to enable end-to-end training of ASA. Specifically, we input the learnable standard deviations into Gradient Reverse Layers (GRL) \cite{ganin2015unsupervised,ganin2016domain} before utilizing them to generate statistics perturbations (see Equ. (\ref{Equ:advstyle}) and Fig. \ref{Fig:advstyle} for more information). By minimizing the task objective solely, the objective minimization w.r.t. the vanilla model parameters and the maximization w.r.t. the learnable standard deviations are achieved simultaneously, thanks to the gradient reverse function of GRL.
	The AdvStyle could be easily implemented and it works in a plug-and-play manner. 
	
	We apply the proposed method on tasks of cross-domain classification and instance retrieval. On standard DG benchmarks,  ASA improves its competitors with higher mean accuracy and lower performance fluctuation.  Especially, a significant performance boost is observed on the PACS dataset under the single source generalization setting (\eg, the performance is boosted from 61.2\% to 67.1\% with a ResNet50 backbone), validating the effectiveness of our proposed ASA method.
	We summarize our contributions as follows:
	\begin{itemize}
		\item We propose a novel Adversarial Style Augmentation (ASA) method, which could explore a broader style space by performing  feature statistics perturbation with less constraints via adversarial training. 
		\item To facilitate the application of ASA in practice, we introduce an AdvStyle module so that ASA can be used in a plug-and-play manner. 
		\item We perform detailed analyses on standard benchmarks of cross-domain classification and instance retrieval. On top of improved mean accuracy, ASA presents lower performance fluctuation, justifying its effectiveness. 
	\end{itemize}
	
	\begin{figure*}[h]
		\begin{center}
			\includegraphics[width=0.78\linewidth]{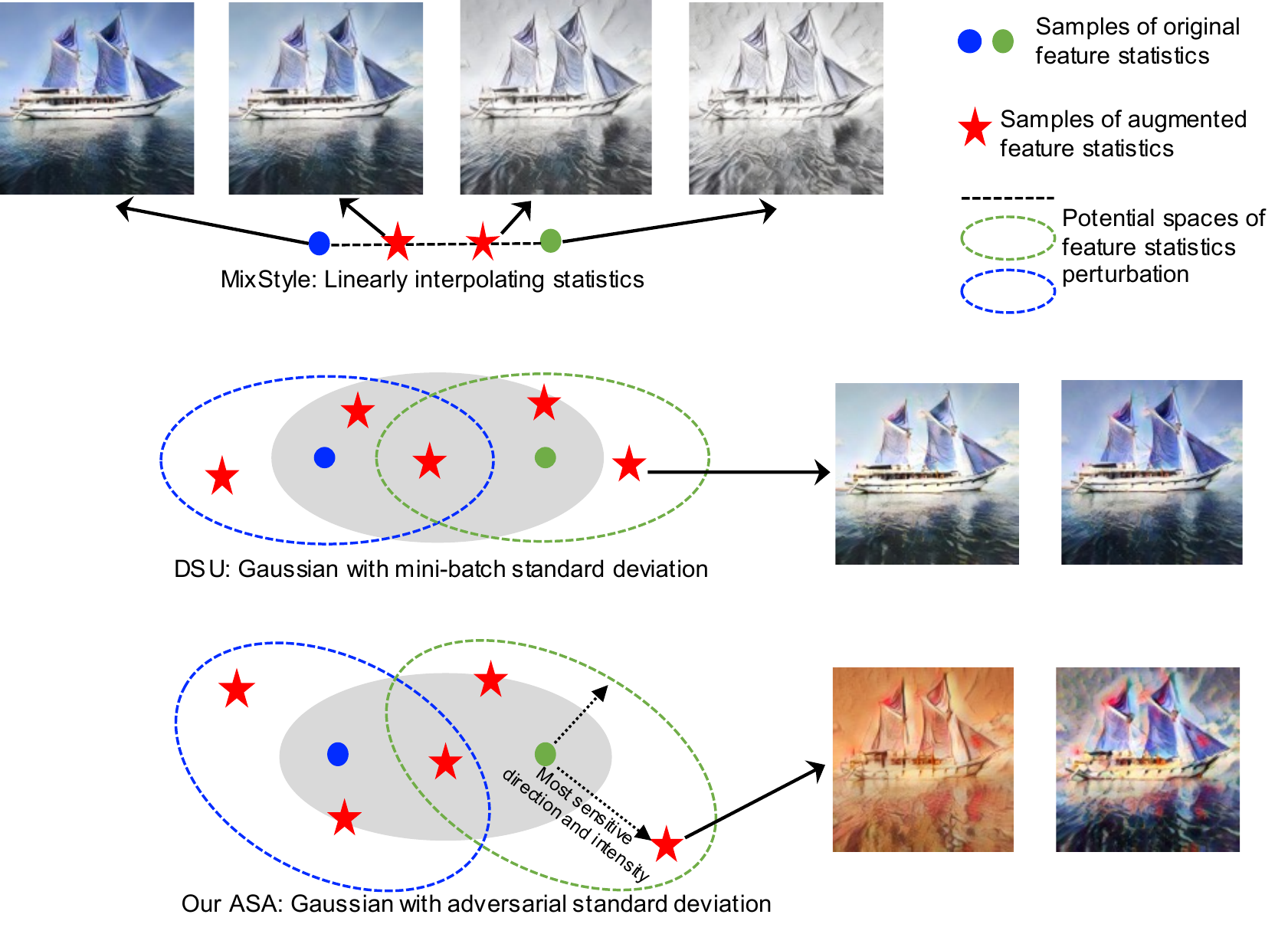}
		\end{center}
		\vspace{-5mm}
		\caption{A toy example on the comparison across different style augmentation-based DG methods. In MixStyle \cite{zhou2021domain}, new styles are introduced by linearly interpolating statistics between two styles. pAdaIN \cite{nuriel2021permuted} is a special case of \cite{zhou2021domain}, where only edge points along the linear connection line are considered via statistics swapping. In \cite{li2022uncertainty}, the uncertainty of  statistics is modeled via multi-variant Gaussian, whose mean and standard deviation are instantiated with the vanilla statistics points and batch standard deviation, respectively. All these methods limit the representation space of statistics within the space spanned by the mini-batch. In contrast, we approach a broader style space by exploring statistics perturbation along the most sensitive direction and intensity that maximize the task objective.
		} \label{Fig:motivations}
	\end{figure*}

	\section{Related Work} \label{Sec:related_work}
	\textbf{Domain generalization.}
	Domain generalization (DG) targets at developing robust DNN models that can perform well on unseen test domains.  The representative DG methods learn domain-invariant feature representations \cite{motiian2017unified,li2018domain,gong2019dlow,zhao2020domain}, or employ meta-learning \cite{li2018learning,du2020learning}, or perform data augmentation \cite{volpi2018generalizing,yue2019domain,zhou2020learning,geirhos2018imagenet,wang2021learning,nam2021reducing,zhou2021domain}.  Our method adopts the data augmentation strategy, more specifically, feature-based augmentation \cite{nuriel2021permuted,zhou2021domain,li2022uncertainty,zhang2022exact}.   
	It has been empirically observed in the task of style transfer \cite{huang2017arbitrary,dumoulin2016learned} that feature statistics can characterize the image styles, and the perturbation on such statistics could yield style-changed but semantic-preserved output. Based on such observation, researchers started to introduce style/distribution augmented training samples by feature statistics perturbation into DG model training \cite{nuriel2021permuted,zhou2021domain,li2022uncertainty,zhang2022exact}. 
	
	To achieve feature statistics perturbation, Nuriel \emph{et al.} \cite{nuriel2021permuted} proposed to swap feature statistics between instances within a mini-batch and, similarly, Zhou \emph{et al.} \cite{zhou2021domain} linearly interpolated feature statistics between instances. Besides the first and second order statistics used in \cite{nuriel2021permuted,zhou2021domain}, Zhang \emph{et al.} \cite{zhang2022exact} implicitly considered high-order statistics for more effective statistics perturbation.  Although improved generalization performance has been observed, the augmented statistics highly rely on the observed feature statistics of training instances, limiting the diversity of statistics. To introduce more diverse statistics perturbations, Li \emph{et al.} \cite{li2022uncertainty} modeled the feature statistics as multi-variate Gaussian distributions and randomly sampled statistics variants from the Gaussians, as illustrated in Fig. \ref{Fig:motivations}. This expands the statistics space of instances, which can be described by the standard deviations of Gaussians. However, these  Gaussian standard deviations are estimated from the mini-batch statistics, still limiting the statistics diversity. 
	
	In this work, we solve the above mentioned problem by acquiring the Gaussian standard deviations with adversarial training. 
	Specifically, instead of estimating Gaussian standard deviations with mini-batch statistics as in \cite{li2022uncertainty}, we model the  Gaussian standard deviations as learnable parameters, leading to a less constrained statistics space. By maximizing the task objective w.r.t. these standard deviations, we explore the most sensitive direction and intensity for statistics perturbation so that the trained DNN model can perform more robustly on unseen test domains. 
	
	\vspace{0.1cm}
	\textbf{Adversarial training.}
	Adversarial training was first introduced in \cite{goodfellow2014generative}, where a discriminator is used to distinguish whether a sample comes from the training data or the generative models. Once the discriminator is fully confused by samples from generators, the generative models successfully recover the training data distribution.
	The adversarial training strategy was later adopted to align feature distributions in domain adaptation \cite{ganin2015unsupervised,ganin2016domain,long2018conditional}, generative photo-realistic super-resolution \cite{ledig2017photo,wang2018esrgan}, data augmentation \cite{zhang2019adversarial,volpi2018adversarial,luo2020adversarial}, and so on.
	Different from most data augmentation methods that introduce augmented samples in the image space \cite{zhang2019adversarial,luo2020adversarial}, we generate augmented samples in the feature space, which is more computationally efficient. 
	Moreover, different from the existing feature augmentation methods \cite{volpi2018adversarial,chen2021adversarial} that conduct adversarial perturbation on raw features, we adversarially change the feature statistics, resulting in specific perturbations along the style dimension. 
	Furthermore, we propose a simple yet effective module to implement our method in a plug-and-play manner, facilitating its usage.


	\section{Methods} \label{Sec:methods}
	Denote by $\vx \in \mathbb{R}^{B\times C\times HW}$ the features encoded by some stacked neural layers, and we respectively denote by $\mu(\vx) \in \mathbb{R}^{B\times C}$ and  $\sigma(\vx) \in \mathbb{R}^{B\times C}$ the mean and standard deviation of features in each channel for an instance, where $B,C,H$ and $W$ represent the batch size, channel dimension, height and width, respectively. Specifically, $\mu(\vx)$ and $\sigma(\vx)$ are computed as:
	\begin{align} \label{Equ:mean_std}
	\mu_{b,c} (\vx) & = \frac{1}{HW} \sum_{i=1}^{HW} \vx_{b,c,i},  \\
	\sigma^2_{b,c}(\vx) & = \frac{1}{HW} \sum_{i=1}^{HW} \left( \vx_{b,c,i} - \mu_{b,c}(\vx) \right)^2
	\end{align}
	where $\mu_{b,c} (\vx)$ represent the mean in the $c$-th channel of the $b$-th instance, and 
	$\sigma^2_{b,c}(\vx)$ and $\vx_{b,c,i}$ are similarly defined. 
	
	It is found in \cite{huang2017arbitrary,dumoulin2016learned,lu2019closed} that the feature statistics, \eg, the mean and standard deviation in Equ. (\ref{Equ:mean_std}) and Equ. (2), can characterize the style/distribution of input images, such as lighting conditions and textures. Therefore, performing statistics perturbation could generate style-changed but semantic-preserved augmented samples, providing an effective method for DG tasks \cite{zhou2021domain,nuriel2021permuted,li2022uncertainty}. Nonetheless, existing methods along this line \cite{zhou2021domain,nuriel2021permuted,li2022uncertainty} mostly utilize feature statistics within the mini-batch to compute statistics perturbation, limiting the statistics representation capacity. In the following, we propose to perform statistics perturbation via adversarial training to overcome this limitation. 
	
	\subsection{Adversarial Style Augmentation} \label{Subsec:adversarial_style_aug}
	Following \cite{zhou2021domain,nuriel2021permuted,li2022uncertainty}, we implement our style augmentation module (SAM) based on the framework of AdaIN \cite{huang2017arbitrary}, which performs style transformation by replacing the feature statistics as:
	\begin{equation}
	\vx_{t} = \sigma_{t} \left( \frac{\vx - \mu(\vx)}{\sigma(\vx)} \right) + \mu_{t},
	\end{equation}
	where $\vx_t$ are the features with new styles decided by $\mu_t \in \mathbb{R}^{B\times C}$ and $\sigma_t \in \mathbb{R}^{B\times C}$. Existing methods typically introduce $\mu_t$ and $\sigma_t$ with feature statistics within a mini-batch, \eg, by randomly shuffling or interpolating feature statistics across instances \cite{zhou2021domain,nuriel2021permuted} or randomly sampling from a Gaussian distribution estimated from the current mini-batch \cite{li2022uncertainty}. Such strategies limit the representation capacity of $\mu_t$ and $\sigma_t$ within a small statistics space. 
	In this paper, we aim to explore larger style spaces by introducing more diverse $\mu_t$ and $\sigma_t$. 
	Specifically, following \cite{li2022uncertainty}, we first model the underlying distribution of $\mu_t$ and $\sigma_t$ as the popular Gaussian with the re-parameterization trick \cite{kingma2013auto}:
	\begin{align} \label{Equ;Gaussian}
	\mu_t  &= \mu(\vx) + \epsilon_{\mu} \Sigma_{\mu}, \quad \epsilon_{\mu}  \sim \mathcal{N}(\mathbf{0},\mathbf{1}), \\
	\sigma_t  &= \sigma(\vx) + \epsilon_{\sigma} \Sigma_{\sigma}, \quad \epsilon_{\sigma}  \sim \mathcal{N}(\mathbf{0},\mathbf{1}),
	\end{align}
	where $\Sigma_{\mu} \in \mathbb{R}^C$ and $\Sigma_{\sigma} \in \mathbb{R}^C$ control the direction and intensity of the statistics perturbation, \ie, the representation space of style augmentation. Therefore, the exploration of style space could be achieved by exploring $\Sigma_{\mu}$ and $\Sigma_{\sigma}$.
	
	\begin{figure}[tb]
		\begin{center}
			\includegraphics[width=0.99\linewidth]{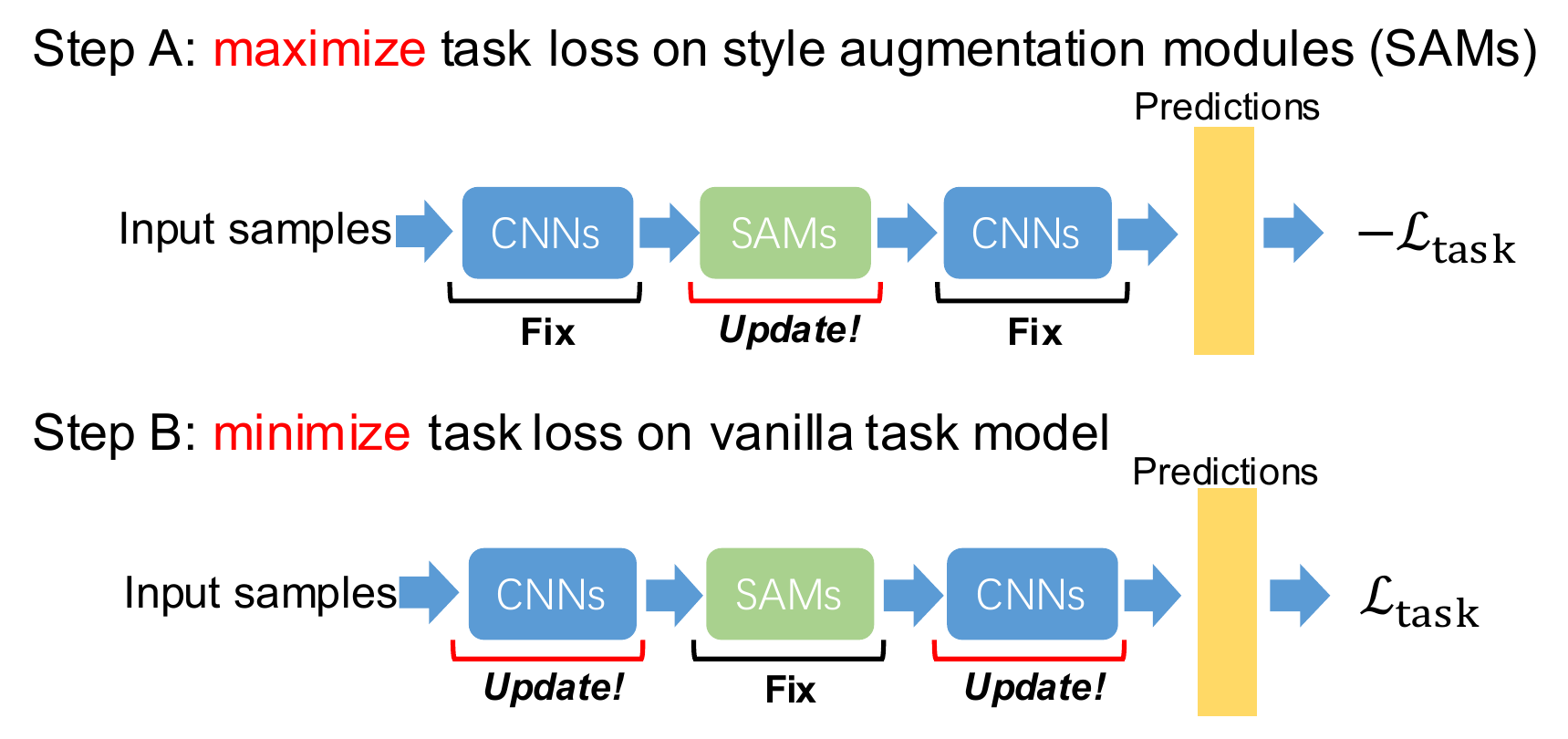}
		\end{center}
		\vspace{-0.3cm}
		\caption{An illustration of the Adversarial Style Augmentation (ASA) method with iterative minimax optimization. Note that we promote to conduct ASA by inserting AdvStyle modules (cf. Sec. \ref{Subsec:advstyle}) in a plug-and-play manner. 
		} \label{Fig:two_step}
	\end{figure}

	Li \emph{et al.} \cite{li2022uncertainty} constructed  $\Sigma_{\mu}$ and $\Sigma_{\sigma}$ as the standard deviations of  $\mu(\vx)$ and $\sigma(\vx)$ along the batch dimension.  Although effective, this method limits the style augmentation space to that of the current mini-batch. 
	To overcome this limitation, we propose to explore a broader style space by imposing less constraints on  $\Sigma_{\mu}$ and $\Sigma_{\sigma}$. 
	
	In \cite{arjovsky2019invariant}, the DG objective is defined as:
	\begin{equation} \label{Equ:dg_objective}
	R^{DG}(f) = \min_{f}\max_{e\in\varepsilon_{all}} R^{e}(f),
	\end{equation}
	where $R^{e}(f)$ refers to the risk of model $f$ within the domain $e$. Given the training domain(s) $\varepsilon_{tr}$, DG aims to learn a model $f$ that performs well across a set of unseen but related domains $\varepsilon_{all} \supset \varepsilon_{tr}$. 
	Assuming that the unknown test domain also belongs to $\varepsilon_{all}$, the objective of Equ. (\ref{Equ:dg_objective}) suggests us to minimize the risk of the worst-case domain among $\varepsilon_{all}$.
	
	Inspired by the DG objective in Equ. (\ref{Equ:dg_objective}), we propose to model $\Sigma_{\mu}$ and $\Sigma_{\sigma}$ as learnable parameters and optimize them via adversarial training, resulting in the following Adversarial Style Augmentation (ASA) method:
	\begin{equation} \label{Equ:overall_loss}
	\min_{\theta}\max_{\Sigma} \mathcal{L}_{task}(\vx, \vy, \theta, \Sigma),
	\end{equation}
	where $\vx$ and $\vy$ are the input data and their corresponding labels, $\theta$ and $\Sigma$ are the parameters of the vanilla task model and the collection of all $\Sigma_{\mu}$ and $\Sigma_{\sigma}$, respectively. $\mathcal{L}_{task}(\cdot)$ is the overall objective defined by the considered tasks. For example, $\mathcal{L}_{task}(\cdot)$ is typically instantiated as the cross-entropy loss in category classification. By learning $\Sigma$ with Equ. (\ref{Equ:overall_loss}), we simultaneously explore the perturbation direction (\eg, the principal orientation direction of $\Sigma$) and intensity (\eg, the Euclidean norm of $\Sigma$), which are individually investigated in Sec. \ref{Subsec:ablation}.

	Let's further clarify the relationship between the DG objective in Equ. (\ref{Equ:dg_objective}) and our proposed ASA objective in Equ. (\ref{Equ:overall_loss}). Since $\Sigma$ controls the representation space of the style perturbation, exploring the worst-case domain of $\varepsilon_{all}$ in Equ. (\ref{Equ:dg_objective}) could be achieved by maximizing the task loss with respect to $\Sigma$ in Equ. (\ref{Equ:overall_loss}). By iteratively exploring the most sensitive $\Sigma$ (\ie, the worst-case domain) for the current model and optimizing the model against such worst-case perturbation, the model is expect to generalize to the worst-case domain, and therefore to any unseen test domains.

	\vspace{-3mm}
	\subsection{AdvStyle} \label{Subsec:advstyle}
	One may opt to implement the ASA method in Equ. (\ref{Equ:overall_loss}) by optimizing $\theta$ and $\Sigma$ iteratively, as presented in Fig. \ref{Fig:two_step}. Here, we propose a simple yet effective module, namely AdvStyle, to instantiate ASA in a plug-and-play manner. 
	Motivated by the seminal adversarial domain adaptation methods \cite{ganin2015unsupervised,long2018conditional}, we propose the AdvStyle module as:
	\begin{equation} \label{Equ:advstyle}
	\vx_{t} = \sigma_{adv} \left( \frac{\vx - \mu(\vx)}{\sigma(\vx)} \right) + \mu_{adv},
	\end{equation}
	where
	\begin{align}
	\mu_{adv}  &= \mu(\vx) + \epsilon_{\mu}  GRL(\Sigma_{\mu}, \lambda), \quad \epsilon_{\mu}  \sim \mathcal{N}(\mathbf{0},\mathbf{1}), \\
	\sigma_{adv}  &= \sigma(\vx) + \epsilon_{\sigma} GRL(\Sigma_{\sigma}, \lambda), \quad \epsilon_{\sigma}  \sim \mathcal{N}(\mathbf{0},\mathbf{1}).
	\end{align}
	where $GRL(\Sigma_{\star}, \lambda)$ is the gradient reverse layer \cite{ganin2015unsupervised}, which outputs the vanilla $\Sigma_{\star}$ in the forward pass and multiplies the gradients with $-\lambda$ in the gradient back-propagation.  $\lambda$ is a hyper-parameter and validated in Sec. \ref{Subsec:ablation}. The gradient reverse layer is widely adopted to perform adversarial training between feature extractor and domain discriminator in domain adaptation \cite{ganin2015unsupervised,long2018conditional}. 
	
	To our best knowledge, we are the first to employ gradient reverse layer to perform adversarial training between style statistics (\ie, $\Sigma$) and the vanilla model (\ie, $\theta$) in DG.  
	Similar to \cite{zhou2021domain,nuriel2021permuted,li2022uncertainty}, we only activate the AdvStyle module for training and deactivate it in the test stage.
	As discussed in Sec. \ref{Subsec:ablation}, conducting ASA by simply inserting AdvStyle into the DNN models gives comparable performance to the iterative optimization-based variant (cf. Fig. \ref{Fig:two_step}). Therefore, we promote to conduct ASA with AdvStyle.

	
	\begin{table*}[h!] \small
		\begin{center}
			\begin{tabular}{l| cccccc}
				\hline
				Method & Art & Cartoon & Photo & Sketch & Mean $\uparrow$ & Std. $\downarrow$\\
				\hline
				ResNet-18   & 58.6$\pm$2.4 & 66.4$\pm$0.7 & 34.0$\pm$1.8 & 27.5$\pm$4.3 & 46.6 & 18.8 \\
				+ pAdaIN \cite{nuriel2021permuted} & 61.5$\pm$2.1 & 71.2$\pm$0.8  & 41.1$\pm$1.9 & 33.1$\pm$3.5& 51.7 & 17.7 \\
				+ MixStyle  \cite{zhou2021domain} & 61.9$\pm$2.2 & 71.5$\pm$0.8 & 41.2$\pm$1.8 & 32.2$\pm$4.1 & 51.7 & 18.1  \\
				+ DSU \cite{li2022uncertainty} & 63.8$\pm$2.0 & 73.6$\pm$0.5 & 39.1$\pm$0.8 & 38.2$\pm$1.2 & 53.7 & 17.8  \\
				+ EFDMix \cite{zhang2022exact} & 63.2$\pm$2.3 & 73.9$\pm$0.7 & 42.5$\pm$1.8 & 38.1$\pm$3.7 & 54.4 & 17.0 \\
				+ AdvStyle  (Ours) & \textbf{67.8}$\pm$0.6 & \textbf{74.5}$\pm$0.4 & \textbf{45.5}$\pm$1.9 & \textbf{47.2}$\pm$1.6 & \textbf{58.7} & \textbf{14.6} \\
				\hline
				\hline
				ResNet-50  & 63.5$\pm$1.3 & 69.2$\pm$1.6    & 38.0$\pm$0.9 & 31.4$\pm$1.5 & 50.5 & 18.3 \\
				+ pAdaIN \cite{nuriel2021permuted} & 67.2$\pm$1.7  & 74.9$\pm$1.4 & 43.3$\pm$0.7 & 36.5$\pm$1.7 & 55.5 & 18.5 \\
				+ MixStyle \cite{zhou2021domain}  & 67.5$\pm$1.5 & 75.2$\pm$1.3 & 42.8$\pm$0.8 & 36.4$\pm$1.2 & 55.5 & 18.8 \\
				+ DSU \cite{li2022uncertainty} & 71.4$\pm$0.2 & 76.9$\pm$1.3 & 42.8$\pm$0.3 & 38.2$\pm$1.1 & 57.3  & 19.6 \\
				+ EFDMix \cite{zhang2022exact} & 75.3$\pm$0.9 & 77.4$\pm$0.8 & 48.0$\pm$0.9 & 44.2$\pm$2.4 & 61.2 & 17.6 \\
				+ AdvStyle (Ours) & \textbf{77.3}$\pm$0.4  & \textbf{78.8}$\pm$0.7 & \textbf{50.3}$\pm$1.5 & \textbf{61.8}$\pm$1.2 & \textbf{67.1} & \textbf{13.6} \\
				\hline
				\hline
				VGG16                                   & 56.2$\pm$0.5 & 62.7$\pm$2.2 & 35.3$\pm$0.7 & 47.5$\pm$1.7 & 50.4 & 11.9 \\
				+ pAdaIN \cite{nuriel2021permuted} & 57.1$\pm$1.1 & 63.7$\pm$1.9 & 36.7$\pm$0.8 & 48.7$\pm$1.6 & 51.6 & 11.7 \\
				+ MixStyle  \cite{zhou2021domain} & 57.3$\pm$0.9 & 64.1$\pm$1.6 & 37.0$\pm$0.6 & 48.6$\pm$1.8 & 51.8 & 11.7  \\
				+ DSU \cite{li2022uncertainty} & 58.3$\pm$1.0 & 65.8$\pm$1.3 & 38.0$\pm$0.4 & 49.7$\pm$2.8 & 53.0 & 11.9 \\
				+ EFDMix \cite{zhang2022exact} & 58.9$\pm$1.1 & 66.2$\pm$0.9 & 38.6$\pm$0.5 & 50.6$\pm$2.3 & 53.6 & 11.8  \\
				+ AdvStyle (Ours) & \textbf{61.9}$\pm$1.0 & \textbf{67.3}$\pm$0.6 & \textbf{40.8}$\pm$0.6 & \textbf{52.9}$\pm$2.5 & \textbf{55.7} & \textbf{11.6} \\
				\hline
				\hline
				JiGen \cite{carlucci2019domain} & 59.7$\pm$1.7 & 67.8$\pm$0.5 & 38.7$\pm$1.7 & 29.0$\pm$3.2 & 48.8 & 18.0 \\
				+ DSU \cite{li2022uncertainty} & 62.6$\pm$1.5 & 72.8$\pm$0.6 & 42.0$\pm$1.4 & 38.3$\pm$2.6 & 53.9 & 16.5  \\
				+ AdvStyle  (Ours) & \textbf{68.2}$\pm$0.9 & \textbf{76.1}$\pm$0.5 & \textbf{48.4}$\pm$1.5 & \textbf{50.8}$\pm$1.3 & \textbf{60.9} & \textbf{13.4}  \\
				\hline
				\hline
				FACT \cite{xu2021fourier} & 69.7$\pm$1.2 & 75.2$\pm$0.4 & 42.7$\pm$1.3 & 48.9$\pm$2.1 & 59.1 & 15.8  \\
				+ DSU \cite{li2022uncertainty} & 72.7$\pm$1.1 & 78.3$\pm$0.3 & 52.9$\pm$1.5 &  62.1$\pm$1.9 & 66.5 & 11.3  \\
				+ AdvStyle  (Ours) & \textbf{74.9}$\pm$0.7 & \textbf{79.1}$\pm$0.4 & \textbf{57.3}$\pm$1.4 & \textbf{67.4}$\pm$1.2 & \textbf{69.7} & \textbf{9.6}   \\
				\hline
			\end{tabular}
			\vspace{-0.2cm}
			\caption{Domain generalization results of classification on the PACS dataset under the single source generalization setting, where the listed domain is adopted for training and results are reported on the remaining three domains.
			} 
			\label{tab:pacs_compre}
		\end{center}
		\vspace{-0.2cm}
	\end{table*}
	
	\section{Experiments}
	
	In this section, we first conduct experiments on tasks of cross-domain classification and instance retrieval to justify the effectiveness of our proposed ASA method, especially the AdvStyle module. 
	Then, ablation studies are provided to analyze the use of our method. Besides, we also justify the effectiveness of our method on the \textit{cross-domain segmentation task} and generalization performance to \textit{images with corruptions} in the \textbf{supplementary material}.
	All experiments are performed under the PyTorch framework on GeForce RTX 2080Ti GPUs.

	\subsection{Generalization on Classification}
	\noindent\textbf{Implementation details.} We conduct the classification experiments on the benchmark PACS dataset \cite{li2017deeper}. There are $9,991$ samples from $7$ classes and $4$ domains, \ie, Art, Cartoon, Photo, and Sketch. We closely follow \cite{zhou2020domain} to prepare the training and test data, set up the optimization strategy, and conduct model selection. Particularly, we perform existing style augmentation-based DG methods \cite{nuriel2021permuted,zhou2021domain,li2022uncertainty,zhang2022exact} under the same setting for fair comparison. 
	Experiments are conducted under the single source generalization setting, where we train models on samples of one domain and test them on the remaining three domains. We also validate the effectiveness of our method in the leave-one-domain-out setting, which is detailed in the \textbf{supplementary material}.
	
	The ResNet-18, ResNet-50 \cite{he2016deep} and VGG16 \cite{simonyan2014very} models pre-trained on the ImageNet dataset \cite{deng2009imagenet} are adopted as the backbones. 
	We also apply our method to existing DG algorithms \cite{carlucci2019domain,xu2021fourier} by inserting the plug-and-play module into their backbones.
	Besides the widely-used mean accuracy across different tasks, we additionally report the standard deviation of classification accuracy across tasks. A smaller standard deviation of accuracy represents smaller performance fluctuation across different tasks, indicating more robust generalization ability.

	\textbf{Results.} All results are shown in Tab. \ref{tab:pacs_compre}. 
	Modelling the uncertainty of statistics via multi-variant Gaussian distribution \cite{li2022uncertainty} typically outperforms those methods based on statistics interpolation \cite{nuriel2021permuted,zhou2021domain} because the statistics space can be more effectively expanded by nonlinear distribution modelling, as we illustrated in Fig. \ref{Fig:motivations} with the toy example.
	By introducing statistics perturbation via adversarial training, we further expand the potential style space towards the worst-case domain, leading to notable performance improvement. For example, by using ResNet-50 as backbone, our method boosts its closest competitor, \ie, DSU, from $57.3\%$ to $67.1\%$ on single source domain generalization, resulting in a $144\%$ relative accuracy improvement over the ResNet-50 baseline (\ie, from $6.8\%$ to $16.6\%$).
	Our method also outperforms the recent work \cite{zhang2022exact} that explores broader style spaces by utilizing high-order batch statistics, revealing the advantage of exploring style spaces beyond batch statistics.
	Additionally, our method achieves the lowest accuracy standard deviation across different tasks with different backbones. This smaller performance fluctuation demonstrates the robust generalization ability of our proposed method.
	
	What's more, it is also found that our method is complementary to existing DG approaches that adopt self-supervised regularization and data augmentation in the image space. For example, Carlucci \emph{et al.} \cite{carlucci2019domain} introduced the self-supervised regularization signals by solving a jigsaw puzzle, while Xu \emph{et al.} \cite{xu2021fourier} proposed a Fourier-based image augmentation strategy to enhance the cross-domain generalization ability. As shown in Tab. \ref{tab:pacs_compre}, by coupling with our proposed AdvStyle, these two methods could be significantly boosted, justifying the nice plug-and-play property of our method.


	\subsection{Generalization on Instance Retrieval}
	We closely follow \cite{zhou2021domain,zhang2022exact} to perform the cross-domain instance retrieval task on person re-identification (re-ID) datasets of Market1501 \cite{zheng2015scalable} and Duke \cite{ristani2016performance,zheng2017unlabeled}. Specifically, we conduct experiments with the OSNet \cite{zhou2019omni} and report the results of ranking accuracy and mean average precision (mAP). As illustrated in Tab. \ref{Tab:reid}, our AdvStyle boosts the vanilla baseline by a large margin (\eg, the mAP is boosted from 25.0 to 32.0 on the Duke$\to$MarKet1501 task). Compared to the common augmentation strategies \cite{zhong2020random,ghiasi2018dropblock}, the style augmentation methods \cite{zhou2021domain,li2022uncertainty,zhang2022exact} present clear advantages. More importantly, our AdvStyle significantly outperforms other style augmentation competitors \cite{zhou2021domain,li2022uncertainty,zhang2022exact}, validating the effectiveness of expanding the style space via adversarial training. 
	
	\begin{table*}[h] \small
		\begin{center}
			\begin{tabular}{L{52.6mm}|C{10.0mm}C{10.0mm}C{10.0mm}C{10.0mm}|C{10.0mm}C{10.0mm}C{10.0mm}C{10.0mm}}
				\hline
				\multirow{2}{*}{Method} & \multicolumn{4}{c|}{MarKet1501$\to$Duke} & \multicolumn{4}{c}{Duke$\to$MarKet1501} \\
				& mAP & R1 & R5 & R10 & mAP & R1 & R5 & R10 \\
				\hline
				OSNet  \cite{zhou2019omni} & 27.9$\pm$0.1 & 48.2$\pm$0.5 & 62.3$\pm$0.1 & 68.0$\pm$0.1 & 25.0$\pm$0.1 & 52.8$\pm$0.8 & 70.5$\pm$0.2 & 77.5$\pm$0.4 \\ 
				+ RandomErase \cite{zhong2020random} & 20.5 & 36.2 & 52.3 & 59.3 & 22.4 & 49.1 & 66.1 & 73.0 \\
				+ DropBlock \cite{ghiasi2018dropblock}  & 23.1 & 41.5 & 56.5 & 62.5 & 21.7 & 48.2 & 65.4 & 71.3 \\
				\hline
				+ MixStyle   \cite{zhou2021domain}  & 28.0$\pm$0.3 & 49.5$\pm$1.1 & 63.6$\pm$0.8  & 68.8$\pm$1.1 & 27.5$\pm$0.2 & 57.4$\pm$0.4 & 74.1$\pm$0.5 & 80.1$\pm$0.1 \\
				+ EFDMix \cite{zhang2022exact} & 29.9$\pm$0.1 & 50.8$\pm$0.1 & 65.0$\pm$0.2 & 70.3$\pm$0.1 & 29.3$\pm$0.3 & 59.5$\pm$0.7 & 76.5$\pm$0.7 & 82.5$\pm$0.1 \\
				+ DSU \cite{li2022uncertainty}  & 31.0$\pm$0.2 & 53.0$\pm$0.3 & 66.6$\pm$0.2 & 71.7$\pm$0.3 & 29.3$\pm$0.2 & 59.9$\pm$0.2 & 77.2$\pm$0.5 & 82.8$\pm$0.4 \\
				+ AdvStyle (Ours) & \textbf{33.2}$\pm$0.1 & \textbf{55.7}$\pm$0.2 & \textbf{69.1}$\pm$0.4 & \textbf{73.5}$\pm$0.5 &   \textbf{32.0}$\pm$0.1 & \textbf{63.2}$\pm$0.1 & \textbf{79.7}$\pm$0.2 & \textbf{85.0}$\pm$0.2 \\
				\hline
			\end{tabular}
			\vspace{-0.1cm}
			\caption{Domain generalization results on the cross-domain person re-ID task.}
			\label{Tab:reid}
		\end{center}
		\vspace{-0.1cm}
	\end{table*}

	\subsection{Ablation and Analyses} \label{Subsec:ablation}
	\begin{table}[h!] 
		\begin{center}
			\begin{tabular}{l| c}
				\hline
				Method & Acc. (\%) \\
				\hline
				\multicolumn{2}{c}{\textbf{Leave-one-domain-out generalization results}} \\
				\hline
				AdvStyle-based ASA    & \textbf{87.0} \\
				Iterative optimization-based ASA  & \textbf{87.0} \\
				\hline
				\hline
				\multicolumn{2}{c}{\textbf{Single source generalization results}} \\
				\hline
				AdvStyle-based ASA   & 67.1 \\
				Iterative optimization-based ASA  & \textbf{68.1}  \\
				\hline
			\end{tabular}
			\caption{Comparisons between the two implementations of the ASA algorithm on PACS dataset.  The test domain and training domain are listed in the leave-one-domain-out setting and single source generalization setting, respectively.} 
			\label{tab:two_asa_imp}
		\end{center}
		\vspace{-0.1cm}
	\end{table}
	
	\noindent\textbf{Implementations of the ASA method.}
	We compare the two implementations of ASA, \ie, the iterative optimization strategy as shown in Fig. \ref{Fig:two_step} and plugging the AdvStyle modules into DNN models. 
	As illustrated in Tab. \ref{tab:two_asa_imp},  the two implementations achieve comparable performance. Though the iterative optimization-based variant presents slightly higher accuracy, we promote the AdvStyle-based variant in practice since it instantiates the ASA in a plug-and-play manner. We adopt the AdvStyle-based variant as the default implementation of ASA in this paper.
	
	\begin{table}[h!] \small
		\centering
		\begin{tabular}{cccccc|c}
			\hline
			Conv-1 & Pool-1 & Res-1 & Res-2 & Res-3 & Res-4 & Acc. (\%) \\
			\hline
			$\checkmark$ & &&&&  & 56.6 \\
			$\checkmark$ & $\checkmark$ & &&&  & 61.9 \\
			$\checkmark$ & $\checkmark$ & $\checkmark$ & &&  & 64.4 \\
			$\checkmark$ & $\checkmark$ & $\checkmark$ & $\checkmark$ & &   & 66.3 \\
			$\checkmark$ & $\checkmark$ & $\checkmark$ & $\checkmark$ & $\checkmark$ &                           & 66.4  \\
			$\checkmark$ & $\checkmark$ & $\checkmark$ & $\checkmark$ & $\checkmark$ & $\checkmark$  &  \textbf{67.1} \\
			\hline
			\multicolumn{6}{c|}{Vanilla ResNet-50 without AdvStyle} &  50.5 \\
			\hline
		\end{tabular}
		\caption{Ablation studies on the inserting position of AdvStyle module on the PACS dataset, where all experiments are based on ResNet-50 backbone and follow the single source generalization setting.  `Conv-1', `Pool-1', `Res-1', `Res-2', `Res-3', `Res-4' denote whether we apply the AdvStyle after the first convolution layer, the first max pooling layer, the first residual block, the second residual block, the third residual block, the fourth residual block, respectively. } 
		\label{Tab:insert_position}
		\vspace{-0.1cm}
	\end{table}
	
	\vspace{0.1cm}
	\textbf{Where to apply AdvStyle.}
	We insert AdvStyle at different positions in the ResNet backbone and present the results in Tab. \ref{Tab:insert_position}. Consistent performance improvements over the vanilla ResNet are observed no matter where the AdvStyle is inserted.  The best performance is achieved when applying the AdvStyle to all the $6$ analyzed positions, which is adopted as the default setting in this paper. 
	
	\begin{figure*}[t]
		\begin{center}
			\includegraphics[width=0.8\linewidth]{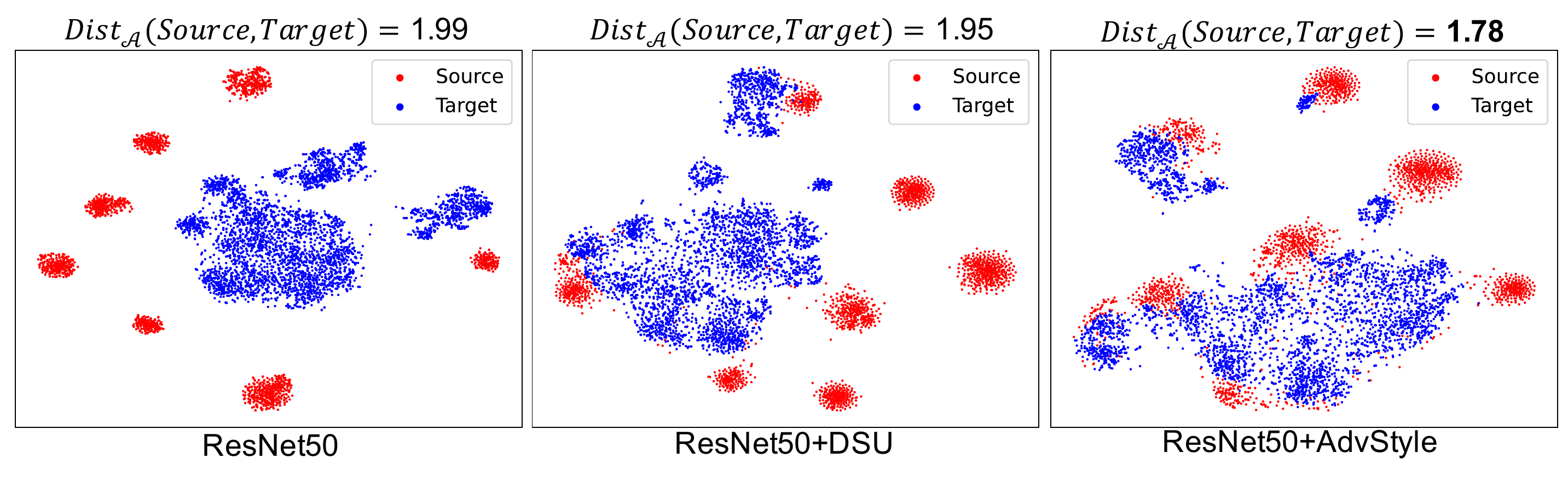} 
		\end{center}
		\vspace{-0.3cm}
		\caption{T-SNE \cite{van2008visualizing} and $\mathcal{A}$-distance (\ie, $Dist_{\mathcal{A}}$) \cite{ben2010theory} of the feature representations on PACS dataset, where a smaller $Dist_{\mathcal{A}}$ represents smaller distribution divergence. We adopt the Art and Sketch as the source and target domains, respectively. More results are provided in the \textbf{supplementary material}.  } \label{Fig:tsne_adis}
	\end{figure*}
	
	\vspace{0.1cm}
	\textbf{The direction and intensity of statistics perturbation.}
	Compared to its closest competitor-DSU \cite{li2022uncertainty}, our AdvStyle introduces different directions (\eg, the principal orientation directions of $\Sigma_{\mu}$ and $\Sigma_{\sigma}$) and intensities (\eg, the Euclidean norms of $\Sigma_{\mu}$ and $\Sigma_{\sigma}$) for statistics perturbation. To investigate the impact of direction and intensity individually, we introduce the AdvStyle-Direction-Only and AdvStyle-Intensity-Only variants.
	In the former case, the perturbation direction is learned via our proposed adversarial training strategy, while the perturbation intensity is set with the batch statistics as DSU \cite{li2022uncertainty}. In the latter case, in contrast, we set the perturbation direction with the batch statistics as DSU \cite{li2022uncertainty} and learn the perturbation intensity via adversarial training. 
	As illustrated in Tab. \ref{tab:direction_intensity}, AdvStyle-Direction-Only and AdvStyle-Intensity-Only  outperform the DSU by 7.6\% and 4.3\%, respectively, indicating that exploring perturbation direction is more effective than exploring the perturbation intensity. The best result is achieved with AdvStyle by exploring both perturbation direction and intensity simultaneously.
	
	\begin{table}[ht] 
		\begin{center}
			\begin{tabular}{l| c}
				\hline
				Method & Acc. (\%) \\
				\hline
				DSU \cite{li2022uncertainty} & 57.3  \\
				AdvStyle-Intensity-Only & 61.6  \\
				AdvStyle-Direction-Only & 64.9  \\
				AdvStyle & \textbf{67.1} \\
				\hline
			\end{tabular}
			\caption{Domain generalization performance on the PACS dataset. Please refer to the main text for the definitions of AdvStyle-Intensity-Only and AdvStyle-Direction-Only.
			} 
			\label{tab:direction_intensity}
		\end{center}
		\vspace{-0.3cm}
	\end{table}

	\textbf{Visualization.} 
	As illustrated in Fig. \ref{Fig:tsne_adis}, a broader style space is successfully achieved by using AdvStyle, which is qualitatively validated by the enlarged overlap regions across domains and quantitatively justified by the reduced $\mathcal{A}$-distance \cite{ben2010theory}.

\begin{figure}[h]
		\begin{center}
			\includegraphics[width=0.8\linewidth]{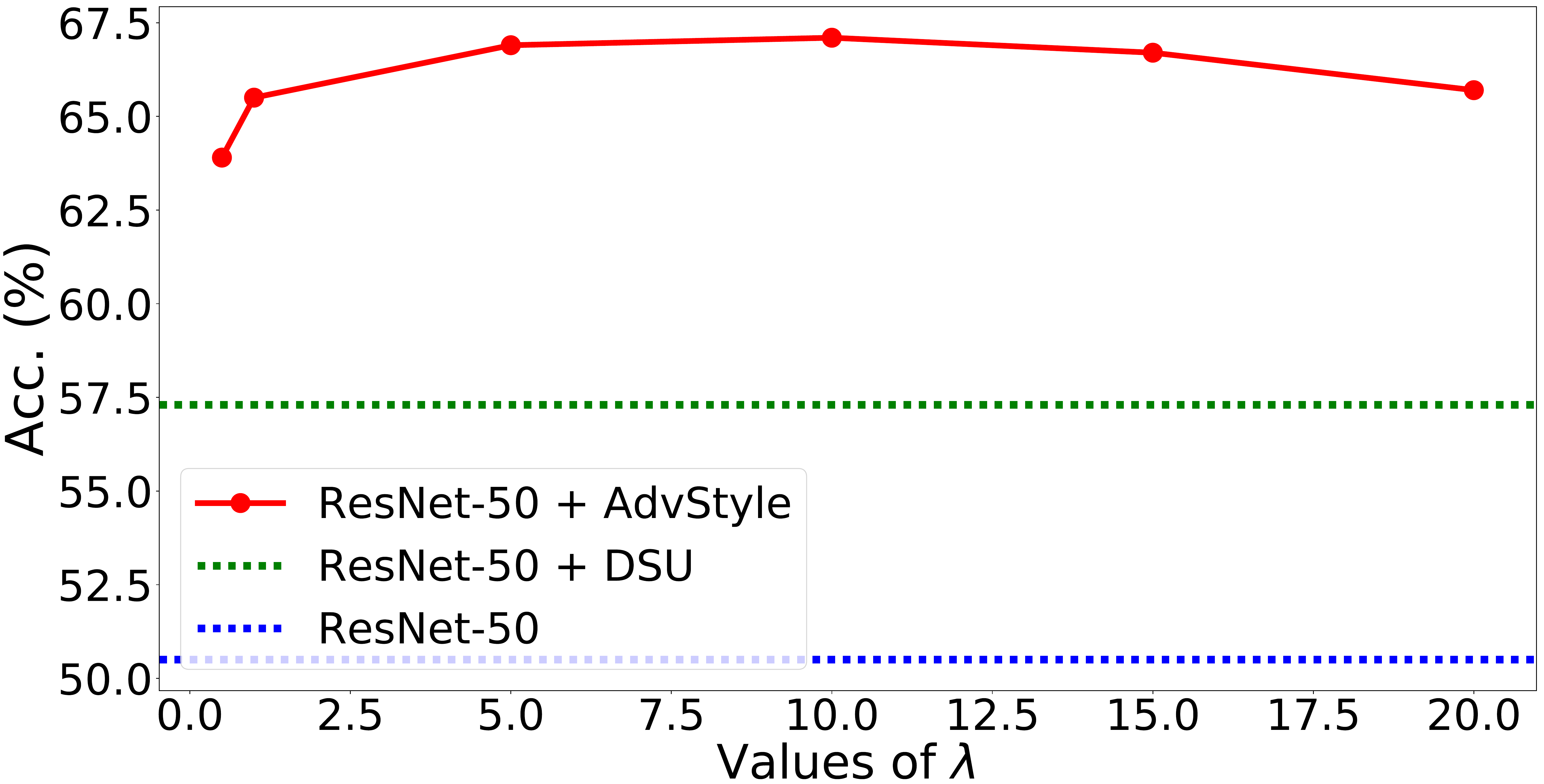} \hfill
		\end{center}
		\vspace{-0.3cm}
		\caption{Results with different values of $\lambda$ on the PACS dataset under the single source generalization setting. 
		} \label{Fig:lambda}
	\end{figure}

	\vspace{0.1cm}
	\textbf{Analyses on the hyper-parameter $\lambda$.} 
	The hyper-parameter $\lambda$ controls the strength of statistics perturbation. 
	As illustrated in Fig. \ref{Fig:lambda}, our method performs stable and significantly outperforms baselines within a large range of $\lambda$ (\eg, $0.5\leq\lambda\leq20$). We empirically find that $\lambda=5$ achieves good results across a large range of tasks, which is adopt the default setting in this paper.

	\vspace{-0.3cm}
	\section{Conclusions}
	To expand the potential statistics space for more diverse style augmentations, we proposed a novel style augmentation method, namely Adversarial Style Augmentation (ASA), by performing feature statistics perturbation via adversarial training. To facilitate the application of ASA in practice, we further introduced a novel module, namely AdvStyle, to instantiate ASA in a plug-and-play manner. ASA outperformed existing style augmentation methods on tasks such as classification and instance retrieval.
	It was also found that expanding style spaces along the direction dimension is more effective than the intensity dimension, which may inspire more studies on the style space exploration.

	%
	%
	\bibliographystyle{splncs04}
	\bibliography{aaai23}
	
\end{document}